\documentclass{article} 
\usepackage{iclr2027_conference,times}
\iclrfinalcopy
\usepackage{soul}

\usepackage[utf8]{inputenc}
\usepackage[T1]{fontenc}
\usepackage{hyperref}
\usepackage{url}
\usepackage{booktabs}
\usepackage{amsmath}
\usepackage{amsfonts}
\usepackage{nicefrac}
\usepackage{microtype}
\usepackage{xcolor}
\usepackage{graphicx}
\usepackage{multirow}

\AtBeginDocument{%
  \setlength{\abovedisplayskip}{5pt plus 1pt minus 2pt}%
  \setlength{\belowdisplayskip}{5pt plus 1pt minus 2pt}%
  \setlength{\abovedisplayshortskip}{3pt}%
  \setlength{\belowdisplayshortskip}{3pt}}

\title{Sketch-Omni: Instance-Level Sketch-Based \\Visual Prompting for Perception MLLMs}

\author{Dongyu Wang, Dar-Yen Chen, Zhiming Hong, Arkaprabha Basu, Yi-Zhe Song\\
SketchX, CVSSP, University of Surrey \\
\texttt{\{dongyu.wang, d.chen, z.hong, a.basu, y.song\}@surrey.ac.uk} \\
}

\let\cite\citep

\begin{document}

\maketitle
\lhead{Preprint. Under review.}
\begin{abstract}
\looseness=-1 Promptable perception, from GroundingDINO to Rex-Omni, puts one model behind one prompt interface and returns coordinates for any localisation task. Every model in that line was built for photographs; none has been given a sketch. We present \textbf{Sketch-Omni}, to our knowledge the first perception MLLM built to work \emph{on} sketches and \emph{from} them. On a sketch, the sketch is the image, and text prompts detect, point, resolve object and spatial referring expressions, and ground affordances in it. From a sketch, the sketch is the prompt, and a drawing of one object localises that object in a photograph, even among several of the same category, a query that neither a category name nor a photo exemplar can pose. Two obstacles stand in the way. A photo-trained model's attention is tuned to colour and texture, which a sketch does not have, and no dataset pairs a drawing with the one instance it depicts among others of its kind. We build the training data instead of collecting it. An edge detector turns every labelled photograph into a sketch-like drawing that keeps the photograph's labels, and two constructions produce sketch-to-photo instance pairs from existing data. On top of this, \emph{photo-guided relation alignment} passes each photograph and its edge sketch through the same model and trains the sketch pass to reproduce how the photograph pass relates phrases to patches and patches to each other. Not one training label is drawn by hand. Sketch-Omni reaches 91.4 F1 on sketch-prompted localisation against 81.8 for the strongest general-purpose MLLM, with a lead that doubles when the photograph holds several instances of the queried category, and it is best in 39 of 40 evaluation cells across six tasks. Code, model, and a hand-annotated benchmark of 2754 sketch-to-photo links will be released.
\end{abstract}

\section{Introduction}

\begin{figure}[t]
\centering
\includegraphics[width=\linewidth]{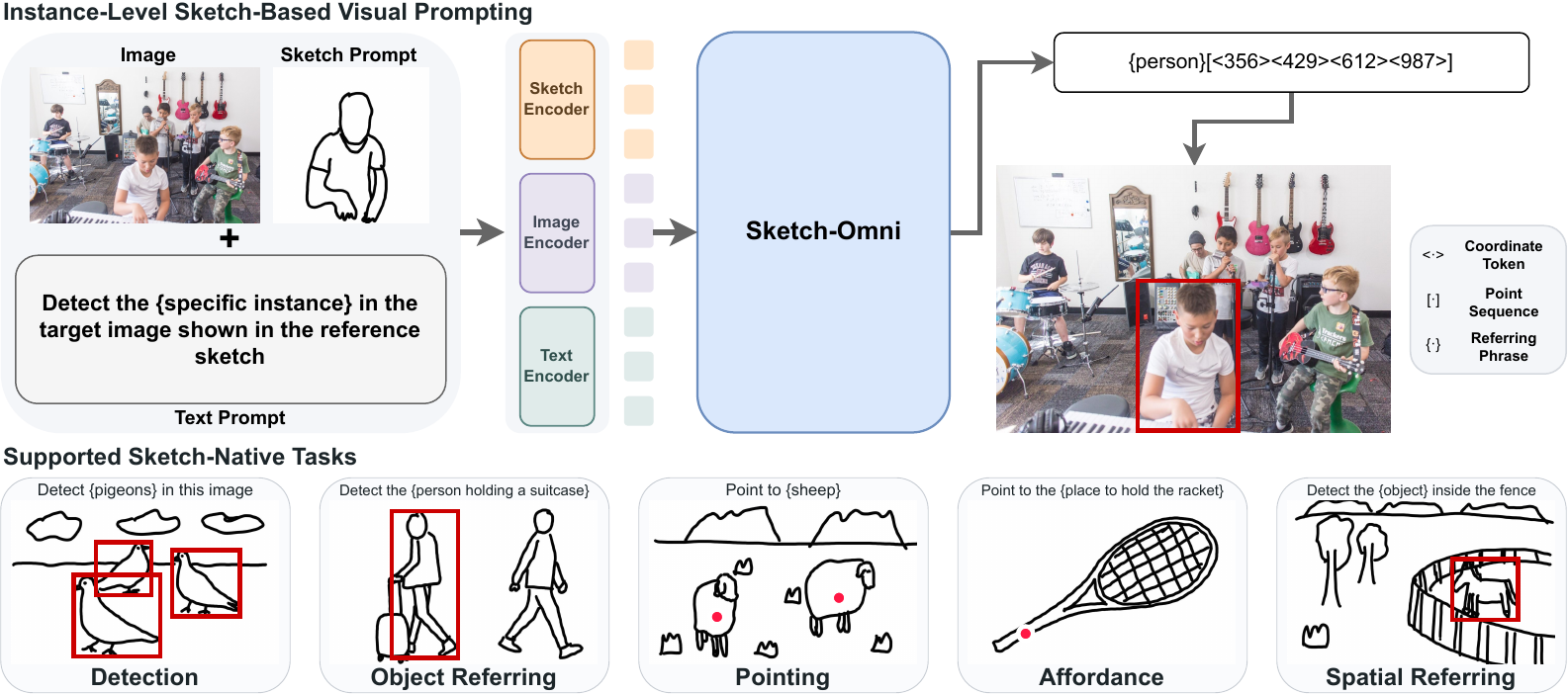}
\caption{\textbf{Sketch-Omni.} Top, \emph{from} a sketch: the sketch is the prompt, and a drawing of one object localises that object in a photograph. Bottom, \emph{on} a sketch: the sketch is the image, and text prompts detect, point, resolve object and spatial referring expressions, and ground affordances in it. One model, one coordinate-token interface.}
\label{fig:teaser}
\end{figure}

Perception has converged on a prompt. GroundingDINO~\cite{groundingdino} opened a detector's vocabulary with a text prompt. T-Rex2~\cite{trex2} let a box or an exemplar stand in for the words. Rex-Omni~\cite{rex_omni} folded detection, pointing, referring, and visual prompting into one MLLM that answers every prompt with coordinate tokens, and general-purpose MLLMs~\cite{qwen25vl,qwen3vl,mimovl,deepseekvl2} now ship grounding as standard. All of these models were built for photographs. None has been given a sketch. This paper brings sketch into the interface, in both directions (Fig.~\ref{fig:teaser}). \textbf{Sketch-Omni} works \emph{on} sketches: the sketch is the image, and text prompts detect, point, resolve object and spatial referring expressions, and ground affordances in it. We call these five the \emph{sketch-native} tasks. And it works \emph{from} sketches: the sketch is the prompt, and given a drawing of one object and a photograph, the model localises that object in the photograph. To our knowledge it is the first perception MLLM to support both directions.

\begin{figure}[t]
\centering
\includegraphics[width=\linewidth]{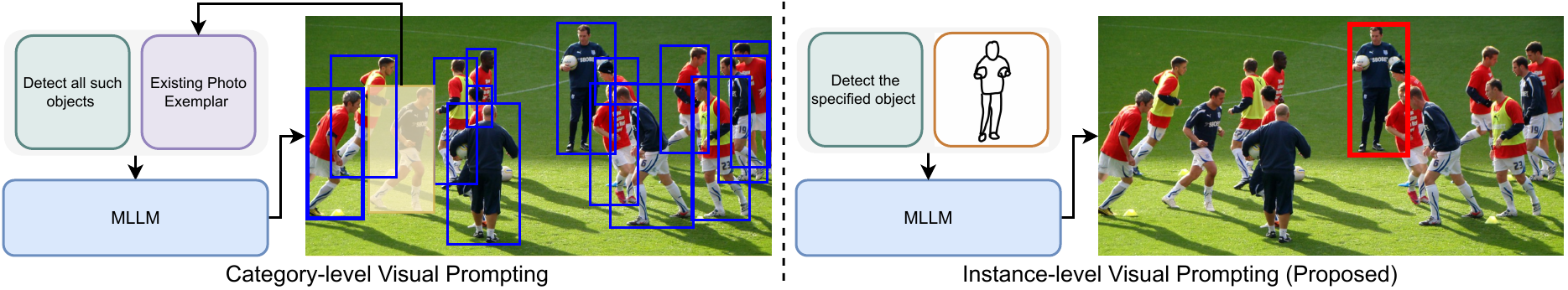}
\caption{\textbf{Category-level versus instance-level visual prompting.} A photo exemplar returns every instance of its category (blue); a sketch returns only the one it depicts (red).}
\label{fig:cat_vs_inst}
\end{figure}

\looseness=-1 Sketch deserves this because it is not another kind of image. A photograph records what stood in front of a camera. A sketch records what a person has in mind: a product not yet built, a suspect never photographed. It is specific: a drawing shows the shape of one particular object, which is why fine-grained sketch retrieval has worked for a decade~\cite{sketch_me_that_shoe,sbir_survey}. And it is cheap: a few strokes, drawn in seconds, before any search begins. Put these together and a sketch can ask a question no other prompt can. A category name returns every instance of the category. A photo exemplar does the same, and it exists only if the object has already been photographed. A sketch, drawn from memory, asks for \emph{that one} among several of its kind (Fig.~\ref{fig:cat_vs_inst}). \citet{ewsod} first showed that a standalone detector can use a sketch this way. No perception MLLM can: its visual prompts are boxes and exemplars, and each stands for a category. We bring the query into the prompt interface as \textbf{instance-level sketch-based visual prompting} (instance VP), and it is where a sketch model should prove itself: on photographs with several instances of the queried category, where a name no longer helps.

\looseness=-1 Two things stand in the way. The first is the model. A photo-trained MLLM grounds a phrase by relating its tokens to image patches, and it relates the patches of one object to each other. Both relations rest on attention tuned to the colour and texture of photographs. A sketch has neither: most of the canvas is blank, and every cue is a line. On a sketch, Rex-Omni's attention spreads over empty space (Fig.~\ref{fig:attn_relation}), and a general-purpose MLLM finds roughly the right region but not the strokes that bound it (Sec.~\ref{sec:exp_sketch_native}). The second is data. Photographic grounding is trained on tens of millions of labelled photographs~\cite{groundingdino,rex_omni}. Sketch has nothing comparable: single-object drawings each paired with one photo~\cite{sketchy}, scene sketches without boxes~\cite{fscoco}, and instance links found by retrieval rather than by hand~\cite{sketchycoco}. Nobody will box-annotate a million drawings.

\looseness=-1 Sketch-Omni answers both with two ingredients. The first is \textbf{supervision built from photographs}. An edge detector~\cite{hed} turns any labelled photograph into a sketch-like line drawing that keeps the photograph's boxes, phrases, and affordance labels; run over Objects365, RefCOCOg, and AGD20K, this supervises all five sketch-native tasks at no labelling cost. For instance VP we construct the pairs no dataset has: a Sketchy drawing whose paired photo is tiled into a canvas beside other photos of its category, and an edge-sketch crop of one annotated object in an Objects365 photograph, to be found again in that photograph. The second is \textbf{photo-guided relation alignment}, which teaches the model to read a sketch the way it already reads a photograph. Grounding rests on two relations: how much each phrase token attends to each image patch, and how the patches of one object relate to each other. Both are about structure, and structure is exactly what a photograph and its edge sketch share. So each labelled photograph passes through the frozen model and its edge sketch through the LoRA-adapted one, and the sketch pass is trained to reproduce the photograph pass's two relations at every layer and head. The features are left free to differ, since a sketch is not a photograph; only the relations are matched. Not one training label is drawn by hand. aThe only hand annotation is for evaluation: 2754 sketch-to-photo instance links on FS-COCO, the first such benchmark we know of.

\looseness=-1 On instance VP, Sketch-Omni reaches 91.4 F1 at IoU 0.5 on FS-COCO, against 81.8 for the strongest general-purpose MLLM, Qwen3-VL-8B, given the same two images; its own base model, whose visual prompts must be boxes inside the target photograph, scores 4.0. The lead over Qwen3-VL-8B is 8.4 F1 when the photograph holds one instance of the queried category and 17.2 when it holds several (Sec.~\ref{sec:exp_stratified}): the advantage sits exactly where a name fails. Across the six tasks Sketch-Omni is best in 39 of 40 table cells, and where general-purpose MLLMs collapse at tight boxes (Qwen3-VL-8B: 75.3 F1 at IoU 0.5 on free-hand detection, 23.0 at IoU 0.9) it holds 65.7. The alignment improves every one of the 14 ablation cells, by up to 4.3 F1 on referring, the task that depends most on a phrase selecting its region (Table~\ref{tab:instance_vp}, App.~\ref{sec:supp_align_native}).

\looseness=-1 \textbf{What this opens.} Sketch research has built one model per task: a recogniser, a retriever, a segmenter. Sketch-Omni replaces the model with a prompt. Any task the grounding literature poses to a photograph can now be posed to a drawing, in the same words; this paper poses six. For grounding research, sketch offers three things: a visual prompt that names an instance without a photograph of it; a modality added to a perception MLLM without a single new label; and a test of how much a model's grounding leans on appearance, since a sketch is a photograph with its appearance removed, and the IoU 0.9 column is the score.

\section{Related Work}
\label{sec:related}

\textbf{Promptable perception.} GroundingDINO~\cite{groundingdino} made a DETR-style detector~\cite{dino} open-set by conditioning it on a text prompt, T-Rex2~\cite{trex2} added visual prompts beside the text, and ChatRex~\cite{chatrex} carried prompted perception into a multimodal LLM. In parallel, Pix2Seq~\cite{pix2seq} cast detection as language modelling over coordinate tokens. Autoregressive coordinate prediction was subsequently adopted by grounding  MLLMs~\cite{kosmos2,shikra,ferret,griffon,cogvlm,florence2} and is now built into general-purpose MLLMs~\cite{qwen25vl,qwen3vl,deepseekvl2,molmo,mimovl,seed15vl}. Rex-Omni~\cite{rex_omni} closes the loop, unifying detection, pointing, referring, OCR, keypoints, visual prompting, and GUI grounding under one coordinate-token vocabulary. Where language cannot name a target, they take clicks, boxes, masks, and regions of interest as prompts~\cite{sam,seem,ferretv2,draw_and_understand,som}, each placed on the image being perceived, so the object is already in view, or taken from another photograph as a category exemplar~\cite{trex2}. The whole line assumes captured images to perceive and photographs to prompt with. Sketch-Omni adds a modality to this line, not a task.

\looseness=-1 \textbf{Sketch as a query.} Sketch research has argued for the drawing as a query since its datasets were built~\cite{tuberlin,sketchy,sketchrnn,sketchycoco,fscoco}, chiefly through sketch-based image retrieval~\cite{sbir_survey,doodle2search,zse_sbir}, and fine-grained SBIR~\cite{sketch_me_that_shoe,sain_cross_modal_hier} established that a drawing identifies an instance rather than a category. Retrieval returns whole images, though. \citet{tripathi_sketch_localization,tripathi_wacv2024} moved the query inside the image, localising sketched objects in natural images, and \citet{ewsod} build a zero-shot detector on a CLIP-initialised~\cite{clip} SBIR backbone that takes the sketch as an instance-aware query. O3SLM~\cite{o3slm} places the sketch query in an LVLM trained on image--sketch--instruction triplets for localisation, counting, retrieval, and question answering. Its sketch stands for a category, since a drawing is paired with any image that holds the drawn class and localisation returns every instance of it, the behaviour of the photo exemplar in Figure~\ref{fig:cat_vs_inst}, and it never perceives within a sketch. The instance-level query therefore exists only in standalone detectors that take no language and serve one task.

\textbf{Perceiving within sketches.} Understanding what a sketch contains has been left to single-task models. Scene sketches have been segmented semantically, recently with an open vocabulary~\cite{openvocab_scene_sketch}, and by instance, where InkLayer~\cite{inklayer} fine-tunes GroundingDINO on scene sketches class-agnostically, collapsing every label to \emph{object}, so the detector survives the domain gap but the prompt does not. SketchFusion~\cite{sketchfusion} fuses foundation-model features into a universal sketch encoder and StructXLIP~\cite{structxlip} aligns CLIP embeddings against structural cues, but none is prompted beyond a category list, and none points, refers, or grounds affordances on a drawing. Sketch-Omni is where the two lines meet.

\section{Sketch-Omni}
\label{sec:method}

Sketch-Omni adds a modality to Rex-Omni~\cite{rex_omni} and leaves its interface unchanged. Stage A (Sec.~\ref{sec:stagea}) transfers the base model's grounding to sketch input through photo-guided relation alignment while training the five sketch-native tasks, and Stage B (Sec.~\ref{sec:stageb}) builds instance VP on that shared reading from constructed supervision.

\looseness=-1 The interface casts perception as next-token prediction. A textual prompt $T$ and the visual inputs form one context, and coordinates are discretised into $1000$ bins, each a dedicated coordinate token (App.~\ref{sec:supp_interface}). \emph{Sketch-native perception} makes the sketch the image input, and the prompt $T$ lists the phrases to be grounded. A LoRA~\cite{lora} update $\Delta\theta$ on the frozen pretrained weights $\theta_0$ defines the \emph{sketch path} and $\theta_0$ alone the \emph{photo path}, written $S$ and $P$ for the input each runs on. Both training stages fit $\Delta\theta$ with a next-token prediction loss $\mathcal{L}_{\mathrm{NTP}}$ on the target sequence, and they interleave under a single optimiser, alternating at a fixed step interval (Sec.~\ref{sec:impl}).

\begin{figure}[tb]
\centering
\includegraphics[width=1\textwidth]{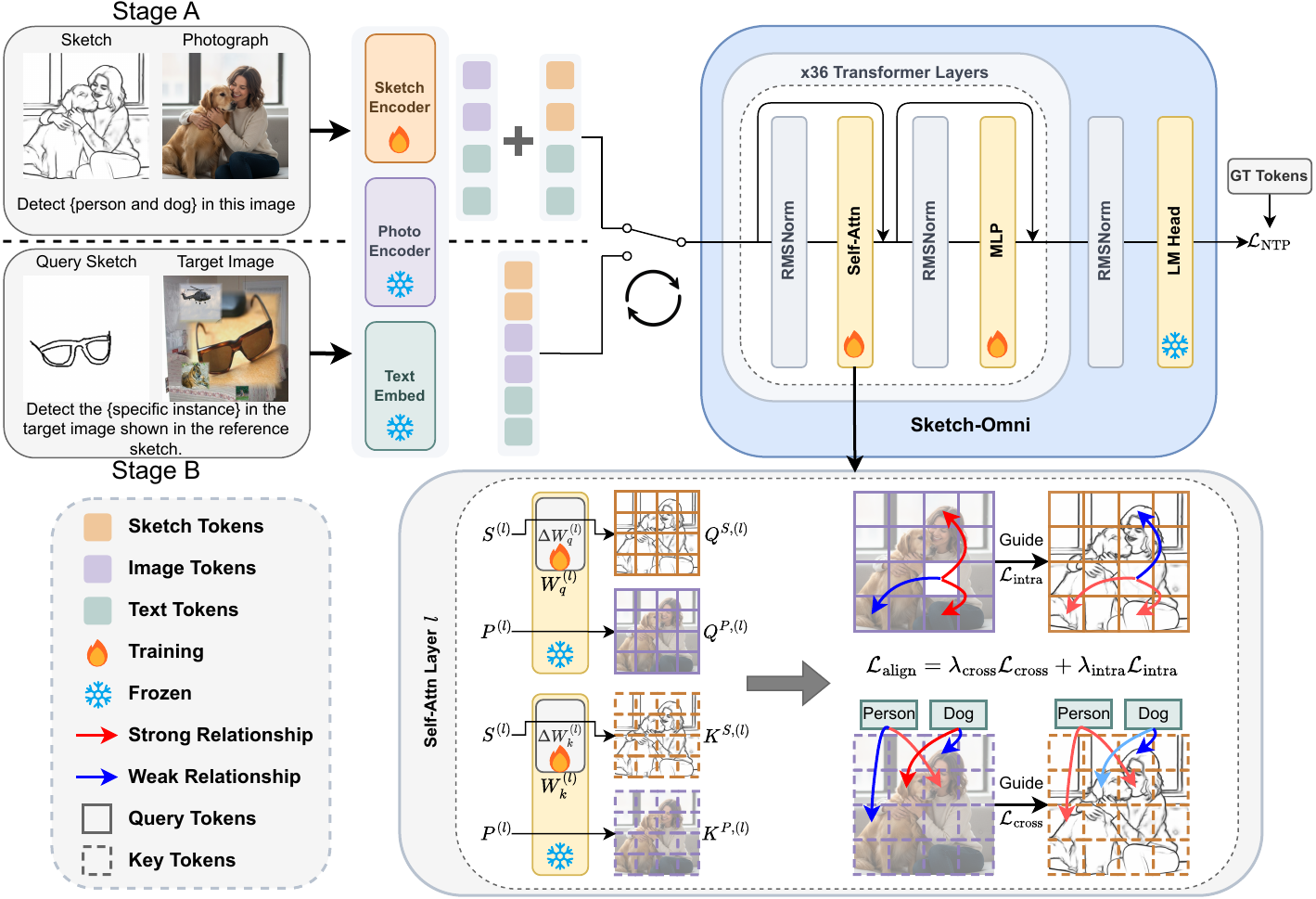}
\caption{\textbf{Sketch-Omni overview.} Left, both stages share one LoRA-adapted backbone under a single optimiser. Right, photo-guided relation alignment matches the sketch path's token relations to the frozen photo path's via $\mathcal{L}_{\mathrm{cross}}$ and $\mathcal{L}_{\mathrm{intra}}$, with solid grids denoting queries and dashed grids keys. $S^{(l)}$ and $P^{(l)}$ are the two paths' hidden states at layer $l$, and $\Delta W^{(l)}_{q}$, $\Delta W^{(l)}_{k}$ the LoRA updates to the frozen projections $W^{(l)}_{q}$, $W^{(l)}_{k}$.}
\label{fig:method}
\end{figure}

\subsection{Stage A: photo-guided relation alignment}
\label{sec:stagea}

\looseness=-1 Stage A trains the five sketch-native tasks, detection, pointing, object referring, spatial referring, and affordance grounding, from two sources. A labelled photograph yields its pixel-aligned edge sketch through a learned detector~\cite{hed}, so the sketch trains under supervision carried over from the photograph. Since a sketch cannot show colour or texture, a lexicon filter~\cite{structxlip} strips appearance terms from each phrase; only the phrases it alters are re-grounded on the photograph by the base model, whose answers become their labels, and every other label stays as annotated. What stays invariant across the two renderings is the scene structure, which patches belong to one object and which region a phrase selects, so the alignment supervises these relations and leaves the features free: it never asks the sketch path to look like the photo path, only to relate its tokens as the photo path relates its own, and the sketch path still encodes strokes.

\paragraph{Which relations to align.}
Grounding rests on two kinds of relation. A phrase has to select its region, and attention already computes this as the cross-modal query-to-key relation, which $\mathcal{L}_{\mathrm{cross}}$ aligns across the two paths. The parts of one object have to hold together, and for this we borrow the intra-modal query-to-query relation underlying DeCLIP's context features~\cite{declip}, which $\mathcal{L}_{\mathrm{intra}}$ aligns in the same way. Both losses compare corresponding token positions between the two readings of one scene, not samples in a batch as in relation-based distillation~\cite{rkd_2019,sp_kd_2019}.

\paragraph{How the relations are matched.}
\looseness=-1 The photo path runs without gradients, so the alignment adds one forward pass per Stage-A step (Fig.~\ref{fig:method}). The sketch keeps the photograph's prompt, target sequence, and patch grid, so a relation in one path has an exact counterpart in the other.

\looseness=-1 At every language-model decoder layer $l$, the sketch path computes $Q^{S,(l)}$ and $K^{S,(l)}$ through the adapted projections and the photo path computes $Q^{P,(l)}$ and $K^{P,(l)}$ through the frozen ones (Fig.~\ref{fig:method}). We align every head of every layer, where each key head is shared by a group of query heads. Cosine similarities are computed after Qwen2.5-VL's rotary position embeddings~\cite{qwen25vl} are applied. Let $\mathcal{V}$ hold the positions of the image tokens, $\mathcal{F} \subseteq \mathcal{V}$ those whose patch overlaps any annotated box, and $\mathcal{T}$ the positions of the phrases in the prompt, all taken on the input side of the teacher-forced sequence. $\mathcal{L}_{\mathrm{cross}}$ keeps every visual position as a key, which lets it penalise phrase similarity that falls on blank regions. $\mathcal{L}_{\mathrm{intra}}$ restricts its queries to $\mathcal{F}$, which keeps the grouping signal on foreground object patches. For simplicity, $q^{S}_i$ and $k^{S}_j$ denote path $S$'s query and key at positions $i$ and $j$, with the layer and head indices suppressed. With $\Phi(u, v) = \langle u, v \rangle / (\|u\| \|v\|)$, $\mathcal{D}$ the decoder layers, and $H$ the number of heads,
\begin{align}
  \mathcal{L}_{\mathrm{cross}} &= \frac{1}{|\mathcal{D}|} \sum_{l \in \mathcal{D}} \sum_{h=1}^{H} \sqrt{\frac{1}{|\mathcal{T}||\mathcal{V}|} \sum_{i \in \mathcal{T},\, j \in \mathcal{V}} \bigl(\Phi(q^{S}_i, k^{S}_j) - \Phi(q^{P}_i, k^{P}_j)\bigr)^{2}},
  \label{eq:align-loss}\\
  \mathcal{L}_{\mathrm{intra}} &= \frac{1}{|\mathcal{D}|} \sum_{l \in \mathcal{D}} \sum_{h=1}^{H} \sqrt{\frac{1}{|\mathcal{F}|^{2}} \sum_{i,\, j \in \mathcal{F}} \bigl(\Phi(q^{S}_i, q^{S}_j) - \Phi(q^{P}_i, q^{P}_j)\bigr)^{2}}.
  \label{eq:intra-loss}
\end{align} With $\mathcal{L}_{\mathrm{align}} = \lambda_{\mathrm{cross}}\, \mathcal{L}_{\mathrm{cross}} + \lambda_{\mathrm{intra}}\, \mathcal{L}_{\mathrm{intra}}$, Stage A optimises
\begin{equation}
  \mathcal{L}_A \;=\; \mathcal{L}_{\mathrm{NTP}} + \mathcal{L}_{\mathrm{align}}
  \label{eq:stagea-loss}
\end{equation}
jointly over the five sketch-native tasks. The two weights are given in Sec.~\ref{sec:impl}.

\subsection{Stage B: instance-level visual prompting}
\label{sec:stageb}

Stage B places two images in the context, a target photograph first and a query sketch second, followed by a prompt naming the sketch as the query (App.~\ref{sec:supp_interface}). The answer is a match between the sketch's tokens and the photograph's, which is what the shared reading built in Stage A makes possible. No dataset annotates the supervision this needs, a free-hand drawing tied to the one instance it depicts inside a larger scene, so we build it in two ways.

\looseness=-1 The first construction uses sketch--photo pair datasets, in which a free-hand drawing is paired with a photograph of a single object. We tile the paired photograph, together with crops of the same category, onto a low-clutter background, so the drawing designates exactly one instance among them, and the category name alone cannot identify the target. The second construction supplies a real scene, with its clutter and occlusion, by deriving pairs from ordinary detection annotations. Given an annotated object, we crop its region from the photograph's edge sketch and use the crop as the query, asking the model to localise that instance in the untouched photograph. One construction trains tolerance to variation in drawing style, the other precise shape matching in real scenes. Full construction details are given in App.~\ref{sec:supp_data}.

When an input holds both a photograph and a sketch, in Stage B and at inference, the vision encoder runs once per image in the interface's fixed image order, with the LoRA update disabled for the photograph and applied for the sketch; the language model applies it over the whole sequence.

Stage B trains with next-token prediction alone, interleaved with Stage A, so the alignment keeps acting on the sketch path while the prompting capability forms on top of it. Alongside the two-image samples, detection and pointing train in this stage on multi-object canvases composited from single-object free-hand drawings, where prompts naming absent categories are answered with an explicit \emph{None}. A Group Relative Policy Optimisation (GRPO)~\cite{grpo} pass with localisation rewards then refines the model (App.~\ref{sec:supp_grpo}), and training sources for both stages are given in Sec.~\ref{sec:exp_datasets}. If shape is what lets a query pick out its instance, the advantage over a competent baseline should widen once same-category distractors are present (Sec.~\ref{sec:exp_stratified}).

\section{Experiments}
\label{sec:exp}

\begin{figure}[!tb]
\centering

\includegraphics[width=1.0\textwidth]{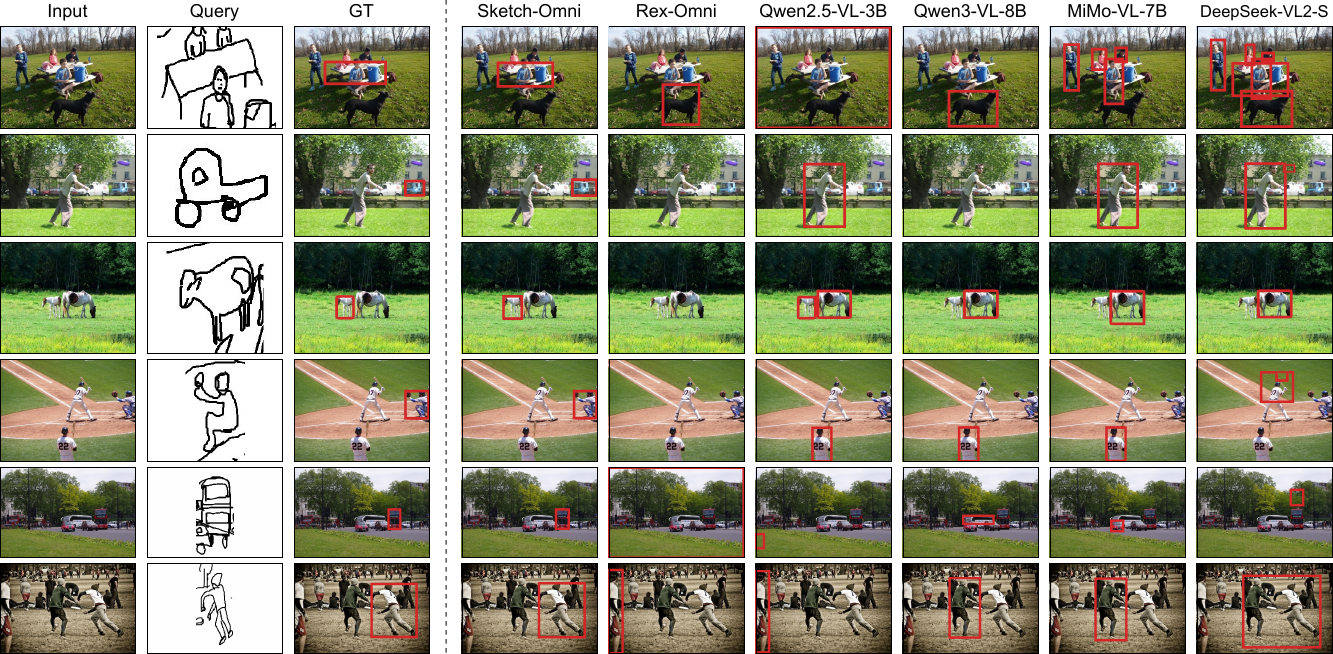}
\caption{\textbf{Qualitative instance-level sketch VP.} Each row shows a free-hand query sketch, the target photograph with the ground-truth box, and one panel per model.}
\label{fig:qual_vp}
\end{figure}

\subsection{Datasets}
\label{sec:exp_datasets}

\paragraph{Stage A, sketch-native perception.} \looseness=-1 Each sketch-native task takes photographs from a photo-grounding dataset and feeds the matching HED edge sketch~\cite{hed} as the input. Objects365~\cite{objects365} supervises detection and pointing, and box centres serve as the pointing targets. RefCOCOg~\cite{refcocog} supervises object referring and spatial referring, which name an instance by what it is or by where it stands, and AGD20K~\cite{agd20k} affordance grounding. 

\paragraph{Stage B, instance-level sketch-based visual prompting.} \looseness=-1 Stage B trains instance-level visual prompting on \textbf{Sketchy}~\cite{sketchy}, which pairs free-hand sketches with single-object photos. One positive crop is tiled alongside same-category distractor crops onto a low-clutter Objects365 background. No Sketchy sketch of a SketchyCOCO foreground category enters this training, and any training sketch that also appears as a SketchyCOCO query is removed. The second construction draws on Objects365~\cite{objects365}, where the query is an annotated object's edge sketch cropped at its box and the target is the source photograph.
\paragraph{Evaluation.} We evaluate against base Rex-Omni~\cite{rex_omni} and baselines appropriate to each task. Detection, referring, and pointing are compared against the general perception MLLMs Qwen2.5-VL~\cite{qwen25vl}, Qwen3-VL~\cite{qwen3vl}, MiMo-VL~\cite{mimovl}, and DeepSeek-VL2~\cite{deepseekvl2}. Instance VP is compared against the same four, where each model receives the query sketch and the target photograph as two images. Since these four MLLMs are not specifically trained on sketches, we also compare against a pipeline that pairs a detector with a sketch retrieval model: GroundingDINO~\cite{groundingdino} proposes boxes in the target photograph under a fixed vocabulary prompt shared by every query, and ZSE-SBIR~\cite{zse_sbir} ranks them against the query sketch. Affordance is compared against the pointing specialists Molmo~\cite{molmo}, RoboPoint~\cite{robopoint}, and RoboRefer~\cite{roborefer}. Referring and affordance are evaluated on HED sketches. These are pixel-aligned with the source photographs, so we reuse the photographs' ground truth. Detection and pointing are evaluated on free-hand sketches from FS-COCO~\cite{fscoco} and on HED sketches rendered from COCO~\cite{coco} under Rex-Omni's protocol.

Free-hand evaluation has no such transfer and requires new labels. SketchyCOCO~\cite{sketchycoco} pairs real photographs with composited scene sketches. However, three properties limit what a margin on it can show: (1) Its queries are linked to their instances by a retrieval model~\cite{sketchy}, so the links are only as accurate as that model. (2) The queries come from a small pool, so the same drawing recurs across different photographs. (3) The queried instance is typically the photograph's dominant object, covering 17.1\% of it at the median against 8.8\% on FS-COCO, so localising the most salient object succeeds without attending to the query. FS-COCO provides what SketchyCOCO cannot, genuine free-hand scene sketches drawn from memory after viewing the paired photograph, but it provides neither object boxes nor instance links. We therefore annotate its test split with boxes on the sketch as ground truth for detection and pointing, and with sketch-to-photo instance links for VP, where each query is the region of the scene sketch that the annotated box covers (App.~\ref{sec:supp_fscoco_annotation}). The evaluation comprises 1340 SketchyCOCO and 2754 FS-COCO sketch queries, with qualitative results in Figure~\ref{fig:qual_vp}.

\paragraph{Metrics.} Given predicted and ground-truth boxes, we match predictions to ground truth one-to-one and greedily by Intersection-over-Union (IoU). Recall and precision are computed over all matches in an evaluation set, and F1 is their harmonic mean. Detection and referring report IoU 0.5, IoU 0.9, and the mean across thresholds from 0.5 to 0.95 with step size 0.05, shown in the \emph{avg} columns. Instance VP reports IoU 0.3, 0.5, and 0.7. Pointing and affordance test containment instead. For pointing, a prediction counts as correct when it falls inside the target's segmentation mask, or its box where no mask exists. For affordance, it must fall inside the ground-truth region obtained by thresholding the heatmap at half its peak (App.~\ref{sec:supp_affordance}).

\begin{figure}[!tb]
\centering
\includegraphics[width=\textwidth]{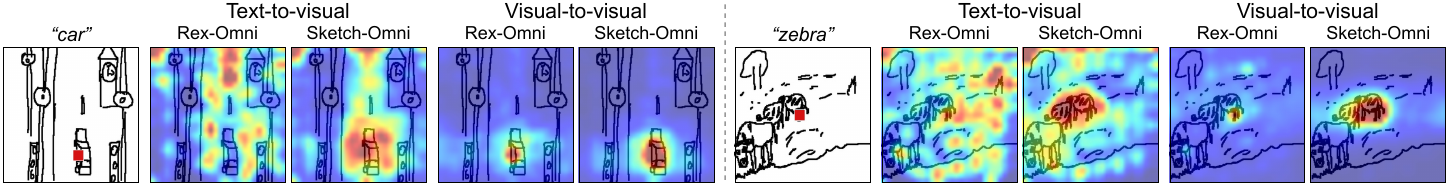}
\caption{\textbf{Cross-modal and intra-modal relations on free-hand sketches.} Cross-modal maps show phrase-query similarity to every visual key, and intra-modal maps the red-marked patch's query against every visual query. All maps are from layer 22.}
\label{fig:attn_relation}
\end{figure}

\subsection{Implementation Details}
\label{sec:impl}

We initialise $\theta_0$ from Rex-Omni~\cite{rex_omni}, a Qwen2.5-VL-3B~\cite{qwen25vl} backbone, and train rank-$64$ LoRA adapters $\Delta\theta$~\cite{lora} on all linear layers of every language-model block and on the query, key, value, and output projections of the vision encoder, with $\theta_0$ frozen. The two stages alternate every $50$ optimiser steps at a Stage-A to Stage-B ratio of $2{:}3$, and the Stage-A objective (Eq.~\ref{eq:stagea-loss}) uses $\lambda_{\mathrm{cross}}{=}0.5$ and $\lambda_{\mathrm{intra}}{=}0.25$. Decoding at inference is greedy. Every configuration, ablations included, is refined by the same final GRPO pass following Rex-Omni. Optimiser and GRPO settings are given in App.~\ref{sec:supp_training}.

\begin{table}[tb]
\centering
\begin{minipage}[t]{0.49\textwidth}
\centering
\scriptsize
\caption{\textbf{Instance-level sketch VP.} Each cell is SketchyCOCO\,/\,FS-COCO, and the lower block ablates the alignment. \textbf{Bold} means best and \underline{underline} means second best.}
\label{tab:instance_vp}
\setlength{\tabcolsep}{4pt}
\renewcommand{\arraystretch}{0.88}
\begin{tabular}{l|ccc}
\toprule
\textbf{Method} & \textbf{F1@0.3} & \textbf{F1@0.5} & \textbf{F1@0.7} \\
\midrule
Qwen2.5-VL-3B & 59.3/55.8 & 53.6/51.1 & 48.8/47.5 \\
Qwen3-VL-8B & 68.3/84.5 & 66.5/81.8 & 62.5/76.9 \\
MiMo-VL-7B & 63.6/48.2 & 61.2/45.5 & 57.2/40.1 \\
DeepSeek-VL2-S & 53.0/43.1 & 50.3/36.0 & 46.5/26.0 \\
Rex-Omni & 18.1/5.7 & 13.0/4.0 & 4.7/1.4 \\
G-DINO\,+\,ZSE-SBIR & 69.5/55.6 & 63.3/51.1 & 57.9/47.1 \\
\midrule
\textbf{Sketch-Omni (full)} & \textbf{74.0}/\textbf{93.6} & \textbf{71.0}/\textbf{91.4} & \textbf{67.7}/\textbf{87.1} \\
\quad w/o align & \underline{73.1}/\underline{91.2} & \underline{69.8}/\underline{88.6} & \underline{66.7}/\underline{84.2} \\
\bottomrule
\end{tabular}
\end{minipage}\hfill
\begin{minipage}[t]{0.47\textwidth}
\centering
\scriptsize
\caption{\textbf{Stratified instance VP.} Each cell is SketchyCOCO\,/\,FS-COCO on the Easy and Hard subsets of Sec.~\ref{sec:exp_stratified}, and the $\Delta$ rows report Sketch-Omni minus Qwen3-VL-8B.}
\label{tab:stratified}
\setlength{\tabcolsep}{3pt}
\renewcommand{\arraystretch}{0.95}
\begin{tabular}{ll|ccc}
\toprule
\textbf{Method} & \textbf{Subset} & \textbf{F1@0.3} & \textbf{F1@0.5} & \textbf{F1@0.7} \\
\midrule
Qwen3-VL-8B & Easy & 91.0/86.9 & \textbf{90.2}/84.3 & \textbf{86.7}/79.5 \\
 & Hard & 50.5/66.2 & 47.8/63.1 & 43.4/56.8 \\
\midrule
\textbf{Sketch-Omni} & Easy & \textbf{91.7}/\textbf{94.8} & 89.5/\textbf{92.7} & 86.1/\textbf{88.3} \\
 & Hard & \textbf{55.3}/\textbf{83.3} & \textbf{51.4}/\textbf{80.3} & \textbf{48.2}/\textbf{77.0} \\
\midrule
\multirow{2}{*}{$\Delta$} & Easy & +0.7/+7.9 & $-$0.7/+8.4 & $-$0.6/+8.8 \\
 & Hard & +4.8/+17.1 & +3.6/+17.2 & +4.8/+20.2 \\
\bottomrule
\end{tabular}
\end{minipage}
\end{table}

\subsection{Instance-Level Sketch-Based Visual Prompting}
\label{sec:exp_vp}

Sketch-Omni outperforms every baseline at all three IoU thresholds (Table~\ref{tab:instance_vp}). Base Rex-Omni fails at this task: its visual prompting takes reference boxes inside the target image and returns every similar object, so it has no instance-level query mode and seldom answers the two-image prompt. Sketch-Omni leads the strongest MLLM by 4.5 and 9.6 F1 at IoU 0.5 on SketchyCOCO and FS-COCO, and the assembled pipeline by 7.7 and 40.3, since its retrieval stage ranks proposals by category and cannot pick one instance out of several. The lower block ablates the alignment. The \emph{w/o align} row is the same recipe trained without $\mathcal{L}_{\mathrm{align}}$,\label{sec:exp_ablation_align} and the constructed supervision alone establishes the capability, outperforming every baseline in every cell. The alignment provides a consistent further improvement of 0.9 to 2.9 points in all six cells. Figure~\ref{fig:attn_relation}\label{sec:exp_ablation}\label{sec:exp_attn} shows how base Rex-Omni and Sketch-Omni differ on a sketch-native detection prompt: cross-modal similarity from the phrase spreads over blank regions under Rex-Omni and concentrates on the queried object under Sketch-Omni, and intra-modal similarity propagates from a token across its instance only under Sketch-Omni. Figure~\ref{fig:qual_vp} shows the failure modes: a baseline that misses returns a same-category distractor, another salient object, or nothing, where Sketch-Omni localises the instance the sketch depicts.

\paragraph{Resolving same-category distractors.}\label{sec:exp_stratified}

\looseness=-1 Instance VP is defined by discrimination among same-category distractors, which single-instance scenes never exercise. We therefore split both benchmarks by the number of annotated instances of the queried category in the photograph, \emph{Easy} when it holds exactly one and \emph{Hard} when it holds two or more. Counts and all MLLM baselines are in App.~\ref{sec:supp_stratified}. Table~\ref{tab:stratified} shows that on FS-COCO-Hard, Sketch-Omni reaches 80.3 F1 at IoU 0.5 against 63.1 for the strongest baseline, 17.2 above it. SketchyCOCO shows the same ordering with a smaller margin, since its same-category instances differ little from one another and leave less for a query to separate. The two models are within one point of each other on SketchyCOCO-Easy. On FS-COCO, whose queries are from the target's own strokes, Sketch-Omni leads by 7.9 to 8.8 F1 even on Easy, and the lead grows with the IoU threshold as a faithful query improves the box itself.

\begin{table}[tb]
\centering
\scriptsize
\caption{\textbf{Sketch-input detection and pointing.} Each cell is FS-COCO\,/\,COCO.}
\label{tab:detection}
\setlength{\tabcolsep}{2pt}
\renewcommand{\arraystretch}{0.95}
\resizebox{\textwidth}{!}{%
\begin{tabular}{l|ccccccccc|c}
\toprule
\multirow{2}{*}{\textbf{Method}} & \multicolumn{9}{c|}{\textbf{Detection}} & \textbf{Pointing} \\
\cmidrule(lr){2-10} \cmidrule(lr){11-11}
 & \textbf{R@0.5} & \textbf{P@0.5} & \textbf{F1@0.5} & \textbf{R@0.9} & \textbf{P@0.9} & \textbf{F1@0.9} & \textbf{R@avg} & \textbf{P@avg} & \textbf{F1@avg} & \textbf{F1@Point} \\
\midrule
Qwen2.5-VL-3B & 48.8/24.2 & 69.7/41.4 & 57.4/30.6 & 13.3/6.5 & 19.1/13.2 & 15.7/8.7 & 30.6/14.8 & 43.7/27.4 & 36.0/19.2 & 72.2/36.3 \\
Qwen3-VL-8B & 70.2/\underline{34.3} & 81.2/\underline{55.6} & 75.3/\underline{42.4} & 21.4/\underline{9.9} & 24.8/17.7 & 23.0/\underline{12.7} & 47.5/\underline{22.0} & 55.0/36.7 & 51.0/\underline{27.5} & 81.7/\textbf{47.9} \\
MiMo-VL-7B & 39.6/19.1 & 43.5/30.2 & 41.5/23.4 & 1.9/2.5 & 2.1/5.6 & 2.0/3.4 & 16.6/8.9 & 18.3/16.3 & 17.4/11.5 & 61.8/33.8 \\
DeepSeek-VL2-S & 57.8/31.5 & 62.3/50.0 & 60.0/38.6 & 11.6/9.3 & 12.5/15.8 & 12.0/11.7 & 33.8/19.9 & 36.5/32.3 & 35.1/24.7 & - \\
Rex-Omni & \underline{74.7}/24.3 & \underline{86.0}/55.1 & \underline{80.0}/33.7 & \underline{52.4}/7.2 & \underline{60.4}/\underline{18.9} & \underline{56.1}/10.4 & \underline{63.5}/15.5 & \underline{73.1}/\underline{37.5} & \underline{68.0}/22.0 & \underline{84.9}/37.9 \\
\midrule
\textbf{Sketch-Omni} & \textbf{81.9}/\textbf{34.5} & \textbf{92.1}/\textbf{79.2} & \textbf{86.7}/\textbf{48.1} & \textbf{62.1}/\textbf{15.8} & \textbf{69.8}/\textbf{37.3} & \textbf{65.7}/\textbf{22.2} & \textbf{72.4}/\textbf{25.7} & \textbf{81.4}/\textbf{59.7} & \textbf{76.6}/\textbf{36.0} & \textbf{89.9}/\underline{44.5} \\
\bottomrule
\end{tabular}%
}
\end{table}

\begin{table}[htbp]
\centering
\begin{minipage}[t]{0.63\textwidth}
\centering
\scriptsize
\caption{\textbf{Sketch-input referring on RefCOCOg.} Each cell is val\,/\,test. Object and spatial referring are told apart by the presence of a spatial term in the expression.}
\label{tab:referring}
\setlength{\tabcolsep}{3pt}
\renewcommand{\arraystretch}{0.95}
\resizebox{\linewidth}{!}{%
\begin{tabular}{l|ccc|ccc}
\toprule
\multirow{2}{*}{\textbf{Method}} & \multicolumn{3}{c|}{\textbf{Object referring}} & \multicolumn{3}{c}{\textbf{Spatial referring}} \\
\cmidrule(lr){2-4} \cmidrule(lr){5-7}
 & \textbf{F1@0.5} & \textbf{F1@0.9} & \textbf{F1@avg} & \textbf{F1@0.5} & \textbf{F1@0.9} & \textbf{F1@avg} \\
\midrule
Qwen2.5-VL-3B & 51.5/50.8 & 15.8/16.7 & 33.7/33.1 & 50.3/50.0 & 13.6/12.5 & 30.2/30.2 \\
Qwen3-VL-8B & 62.2/62.4 & 23.7/25.5 & 43.8/44.3 & 61.7/62.5 & \underline{21.1}/\underline{21.9} & 42.3/42.8 \\
MiMo-VL-7B & 43.5/45.7 & 2.3/3.4 & 20.0/22.4 & 41.7/42.7 & 1.8/2.4 & 18.4/18.6 \\
DeepSeek-VL2-S & \underline{69.1}/\underline{66.1} & \underline{25.4}/\underline{25.9} & \underline{48.2}/\underline{46.5} & \underline{67.6}/\underline{65.1} & \underline{21.1}/20.2 & \underline{43.9}/\underline{43.0} \\
Rex-Omni & 52.0/51.2 & 20.2/20.3 & 36.3/35.7 & 54.2/54.4 & 19.4/19.8 & 37.1/37.5 \\
\midrule
\textbf{Sketch-Omni} & \textbf{75.0/74.6} & \textbf{43.1/42.7} & \textbf{59.6/59.6} & \textbf{77.6/76.7} & \textbf{43.3/43.1} & \textbf{59.9/60.7} \\
\bottomrule
\end{tabular}}
\end{minipage}\hfill
\begin{minipage}[t]{0.35\textwidth}
\centering
\scriptsize
\caption{\textbf{Affordance grounding on AGD20K.} Point-in-region F1 on the Seen and Unseen splits, against pointing specialists.}
\label{tab:affordance}
\setlength{\tabcolsep}{4pt}
\renewcommand{\arraystretch}{0.95}
\begin{tabular}{lcc}
\toprule
\textbf{Method} & \textbf{Seen} & \textbf{Unseen} \\
\midrule
Molmo-7B-D & 53.6 & 56.8 \\
Molmo-72B & \underline{56.8} & 58.6 \\
RoboPoint & 13.8 & 13.9 \\
RoboRefer-2B & 40.6 & 43.7 \\
Rex-Omni & 51.3 & \underline{61.0} \\
\midrule
\textbf{Sketch-Omni} & \textbf{66.2} & \textbf{74.3} \\
\bottomrule
\end{tabular}
\end{minipage}
\end{table}

\begin{figure}[!tb]
\centering

\includegraphics[width=1\textwidth]{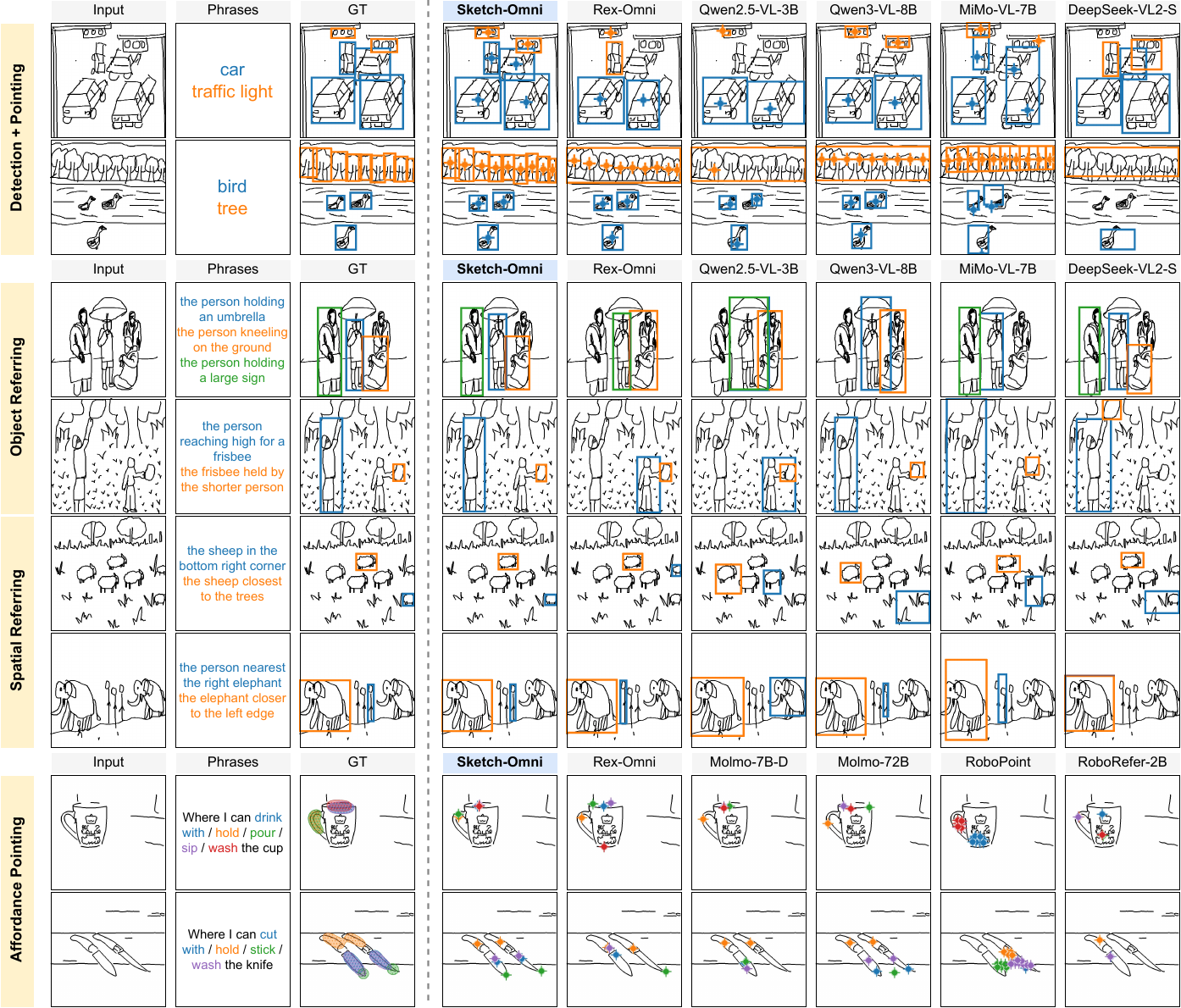}
\caption{\textbf{Qualitative results on sketch input.} Two examples each of detection and pointing, object referring, spatial referring, and affordance grounding, with box and point colours matching the phrase colours. The referring expressions and their ground-truth boxes are newly annotated for these examples, and DeepSeek-VL2-S outputs boxes only.}
\label{fig:qualitative}
\end{figure}

\raggedbottom
\subsection{Sketch-Native Tasks}
\label{sec:exp_sketch_native}

Sketch-Omni achieves the best recall, precision, and F1 in every detection column of Table~\ref{tab:detection}, with qualitative results in Figure~\ref{fig:qualitative}. On free-hand sketches, Sketch-Omni attains its largest margin at F1@0.9, leading base Rex-Omni by 9.6 and the best general-purpose MLLM by 42.7. Sketch-Omni retains 62.1 of its 81.9 recall as the threshold tightens from 0.5 to 0.9, while Qwen3-VL drops from 70.2 to 21.4, the signature of grounding tuned to appearance: the region is found and the strokes are not. On edge sketches, Sketch-Omni leads the best baseline by 23.6 in precision at IoU 0.5. On pointing, Sketch-Omni outperforms every model on free-hand sketches. 

On object and spatial referring (Table~\ref{tab:referring}), Sketch-Omni achieves the best result in every cell, and attains its largest margin at IoU 0.9, where object F1 reaches 43.1/42.7 against 25.4/25.9 for the strongest baseline. Every general-purpose MLLM loses average F1 on spatial expressions relative to its own object subset, up to 4.3 points for the strongest, while Sketch-Omni widens its margin from 11.4/13.1 to 16.0/17.7 and resolves spatial expressions better than object ones, 77.6 against 75.0 F1 at IoU 0.5 on val, which is what a drawing that keeps layout and drops appearance should do.

\looseness=-1 On AGD20K affordance grounding (Table~\ref{tab:affordance}), the 3B Sketch-Omni outperforms every pointing specialist on both splits, improving base Rex-Omni by 14.9 and 13.3. The alignment accounts for 0.4 to 4.3 F1 of the gain across the five tasks (App.~\ref{sec:supp_align_native}). Across Tables~\ref{tab:instance_vp}, \ref{tab:detection}, \ref{tab:referring}, and~\ref{tab:affordance}, Sketch-Omni is best in 39 of 40 cells.

\section{Conclusion}
\label{sec:conclusion}

Sketch-Omni brings promptable perception to sketch, on the drawing and from it. One diagnosis, that a sketch keeps structure and drops appearance, gives the recipe: construct the supervision, and align relations rather than features, with no new training label. The evidence concentrates where a name fails, among several instances of one category, and there Sketch-Omni leads the strongest baseline by 3.6 F1 on SketchyCOCO and 17.2 on FS-COCO at IoU 0.5. For sketch research, every perception task is now a prompt away; for grounding, sketch is a prompt that carries identity, a modality that costs no labels, and a test of what grounding leans on.

\subsection*{Acknowledgements}
The authors acknowledge the use of resources provided by the Isambard-AI National
AI Research Resource (AIRR). Isambard-AI~\cite{isambard_ai} is operated by the
University of Bristol and is funded by the UK Government's Department for
Business, Innovation, Science and Trade (DBIST) via UK Research and Innovation;
and the Science and Technology Facilities Council [ST/AIRR/I-A-I/1023].

\bibliography{references}
\bibliographystyle{iclr2027_conference}

\clearpage
\appendix
\section*{Appendix}
\section{Token Interface of the Base Model}
\label{sec:supp_interface}

Rex-Omni~\cite{rex_omni} casts every perception task as conditional sequence generation over one shared vocabulary, so detection, pointing, referring, and visual prompting differ only in their prompt and output content.

\paragraph{Input grammar.}
A visual input is encoded by the vision encoder into $N_v$ patch tokens under Qwen2.5-VL's dynamic-resolution convention, so $N_v$ varies with the input's size and aspect ratio. The patch tokens are wrapped in the image delimiters $\langle\text{im}\rangle$ and $\langle/\text{im}\rangle$, then concatenated with the $N_t$ tokens of the textual prompt $T$,
\begin{equation}
  \mathbf{x} \;=\; \big[\, \langle\text{im}\rangle,\, v_1,\, \ldots,\, v_{N_v},\, \langle/\text{im}\rangle,\, w_1,\, \ldots,\, w_{N_t} \,\big].
  \label{eq:input-grammar}
\end{equation}
For sketch-native tasks $T$ carries phrases $p_1, \ldots, p_M$, as in ``Detect $p_1, \ldots, p_M$ in the sketch.'', where $M$ is the number of phrases and a phrase may be a category name, a referring expression, or an affordance verb. Sketch VP uses a fixed prompt, since the query is carried by the sketch span rather than by words. The position sets that Sec.~\ref{sec:method} pairs index this sequence: $\mathcal{V}$ holds the positions of $v_1, \ldots, v_{N_v}$, $\mathcal{T}$ those of the tokens of $p_1, \ldots, p_M$ within $w_1, \ldots, w_{N_t}$, and $\mathcal{F} \subseteq \mathcal{V}$ those whose patch overlaps an annotated box.

\paragraph{Output grammar.}
Coordinates are normalised to the image and quantised into $1000$ bins, each a dedicated
\emph{coordinate token} in the vocabulary. Every task answers with one structured token sequence,
built from a descriptive phrase, coordinate tokens, and special tokens that demarcate the two,
\begin{equation}
  \underbrace{\langle\text{ref}\rangle\, \textsc{phrase} \,\langle/\text{ref}\rangle}_{\text{phrase}} \;\; \underbrace{\langle\text{box}\rangle\, \textsc{coords} \,\langle/\text{box}\rangle}_{\text{geometry}},
  \label{eq:io-grammar}
\end{equation}
where \textsc{phrase} names the object the coordinates stand for, and \textsc{coords} lists them,
$[x_0, y_0, x_1, y_1]$ per box and $[x_0, y_0]$ per point. Several instances of one phrase share a
single geometry segment, comma-separated and ordered by increasing $x_0$, and commas likewise join
the blocks of different phrases. A phrase with no match in the input carries a literal \emph{None} inside the geometry segment.

\paragraph{Visual prompting in the base model.}
Rex-Omni prompts visually with boxes rather than with a second image. One or more reference boxes inside the target image are quantised into the same coordinate tokens as an output box and passed in the prompt as a list under a single object key, and the released template reads ``Given reference boxes \{``object\_1'': [``$c_1 c_2 c_3 c_4$'', \ldots]\} indicating one or more objects, find all similar objects in the image and output their bounding boxes''. The answer lists every box of the indicated category. The query therefore names a category by example and lives in the image it queries, which is why it has no way to ask for one instance among several of that category (Sec.~\ref{sec:exp_ablation}).

\paragraph{Sketch-VP extension.}
The sketch-VP input extends Eq.~\ref{eq:input-grammar} to two delimited visual spans, the target photograph first and the query sketch second, with the prompt specifying the query sketch,
\begin{equation}
  \mathbf{x}_{\text{vp}} \;=\; \big[\, \langle\text{im}\rangle\, v^{P}_1,\ \ldots,\ v^{P}_{N_P},\ \langle/\text{im}\rangle,\, \langle\text{im}\rangle\, v^{S}_1,\ \ldots,\ v^{S}_{N_S},\ \langle/\text{im}\rangle,\, w_1,\, \ldots,\, w_{N_t} \,\big],
  \label{eq:input-grammar-vp}
\end{equation} where $v^{P}_1, \ldots, v^{P}_{N_P}$ are the patch tokens of the target photograph and $v^{S}_1, \ldots, v^{S}_{N_S}$ those of the query sketch. The output grammar is Eq.~\ref{eq:io-grammar} unchanged, with the queried instance's category name filling the reference segment and the answer box expressed in the target photograph's coordinate frame. Sketch-Omni therefore inherits Rex-Omni's decoding, parsing, and evaluation stack verbatim.

\section{Training Data Construction}
\label{sec:supp_data}

\begin{figure}[!htbp]
\centering
\includegraphics[width=\textwidth]{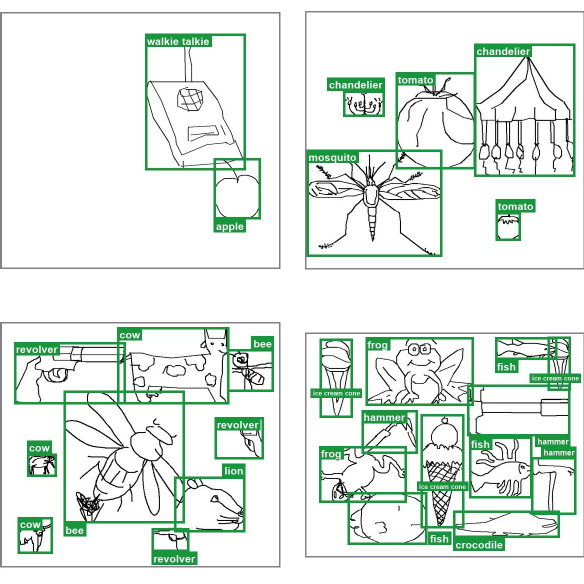}
\caption{\textbf{Composited free-hand sketch scenes.} Four scenes with their ground-truth boxes, ordered by instance count. Canvas size and aspect ratio are sampled per scene.}
\label{fig:constr_scene}
\end{figure}

\paragraph{Composited sketch scenes for Stage B detection and pointing.}
The scenes composite single-object drawings from three sources, each contributing a different kind. QuickDraw~\cite{sketchrnn} holds 345 categories of doodles drawn under a twenty-second limit, TU-Berlin~\cite{tuberlin} 250 categories drawn under a much looser thirty-minute limit, and Sketchy~\cite{sketchy} 125 categories drawn from memory after a brief view of a photograph, so mixing them exposes Stage B to the range of abstraction a free-hand query can take. All three store strokes as vectors. Their class names are unified into a self-derived vocabulary of 415 categories, the canonicalised union of the three: names are lowercased, de-pluralised, and passed through a synonym table that merges variants such as \emph{aeroplane} and \emph{airplane}. For each instance, a source dataset is first sampled uniformly, then a category within that source, then a drawing, so the mix stays near one third per source and QuickDraw's far larger pool cannot dominate. Each scene holds 2 to 12 instances from 2 to 6 categories on a canvas whose long side varies from 768 to 1152 pixels (Fig.~\ref{fig:constr_scene}). Every drawing is rendered from its stroke data directly at its final canvas size rather than rescaled, so stroke width stays constant across object sizes. A placement is rejected when the new box overlaps a placed one above $0.5$ IoU, or when either box contains more than $0.8$ of the other, so no object ends up hidden inside another. Prompts list the present categories together with sampled absent ones, which the target answers with \emph{None}. Scenes are generated procedurally during training rather than stored.

\begin{figure}[!htbp]
\centering
\includegraphics[width=\textwidth]{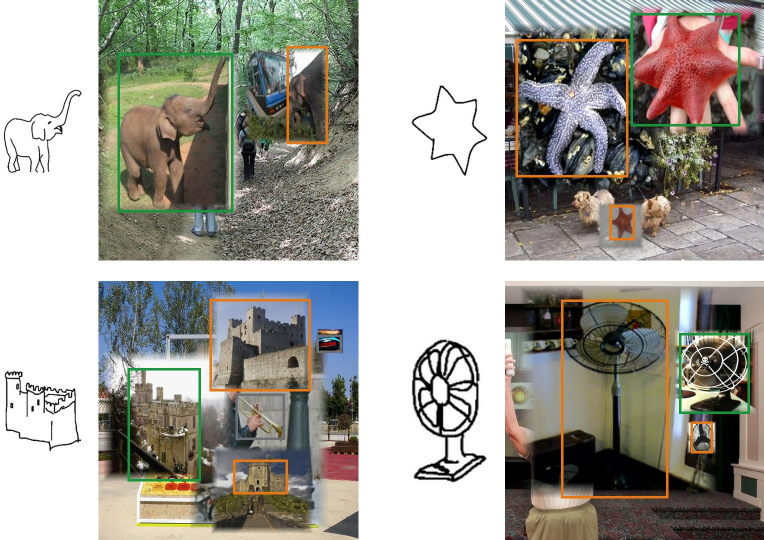}
\caption{\textbf{Stitched Sketchy canvases.} Each group holds the query sketch and the canvas it is asked about. Green marks the queried instance, orange a same-category distractor, and grey every other object.}
\label{fig:constr_sketchy}
\end{figure}

\paragraph{Sketchy stitched canvases for Stage B.}
Sketchy provides photo-sketch pairs but no bounding boxes, so each photo receives a tight box from GroundingDINO~\cite{groundingdino} prompted with its category name. A square canvas with side sampled between 768 and 1280 pixels tiles 1 to 7 such photos drawn from up to 4 categories (Fig.~\ref{fig:constr_sketchy}). Instance-level samples place several tiles of the queried category, so the query competes with a same-category distractor. Tiles are cropped off-centre within the margin around the tight box, resized across a stratified size range, colour jittered, alpha-feathered at their borders, and placed so that no tile overlaps a previously placed object region. Every ground-truth box therefore stays fully visible. Backgrounds are Objects365 photographs screened to contain none of the pasted categories. Query sketches are inverted to white-on-black so their polarity matches the HED sketches of Stage A.

\paragraph{Objects365 HED-sketch queries for Stage B.}

\begin{figure}[!htbp]
\centering
\includegraphics[width=\textwidth]{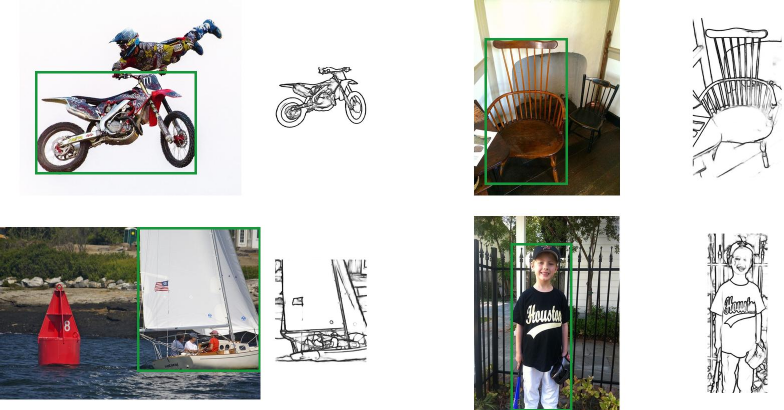}
\caption{\textbf{Objects365 HED-sketch queries.} Each group holds a photograph with the queried box and the HED-sketch crop taken from that box, which serves as the query. Sketches are drawn black on white here, and training uses the inverted polarity.}
\label{fig:constr_obj365}
\end{figure}

For the Objects365 augmentation, HED is applied to the whole photograph, and the query is cropped from the resulting edge sketch at one annotated object's box, with a random padding of 2.5 to 7.5\% of the box size per side. Boxes with area below $96\times96$ pixels are skipped. The model is then asked to localise that instance in the untouched source photograph (Fig.~\ref{fig:constr_obj365}).

\paragraph{SketchyCOCO overlap filter.}
SketchyCOCO's foreground sketches were themselves retrieved from QuickDraw, TU-Berlin and Sketchy, the three pools the constructions above draw from, so an evaluation query could otherwise appear among the training drawings. Two levels of filtering separate them. At the category level, every Sketchy pair whose category is one of SketchyCOCO's 14 foreground categories is dropped from training: airplane, bicycle, car, cat, cow, dog, elephant, fire hydrant, giraffe, horse, motorcycle, sheep, traffic light and zebra. At the instance level, every SketchyCOCO evaluation query is matched against all three pools and every drawing it matches is denied to both constructions. A drawing's signature is its ink cropped to the tight box, stretched to $32 \times 32$, Gaussian blurred and L2 normalised, and a candidate is denied once its cosine similarity to a query reaches $0.92$, with the horizontal flip of each query matched as well. The stretch removes the aspect distortion SketchyCOCO applies when fitting a sketch into an instance box, and the blur absorbs the stroke-width and antialiasing differences between its renders and ours.

\section{FS-COCO Annotation}
\label{sec:supp_fscoco_annotation}

\paragraph{Procedure.} Two people annotated the official FS-COCO test split and a third checked every annotation, using a browser tool built for the purpose. It shows the photograph, with COCO's boxes drawn on it, beside the rendered sketch, and records every box and link as it is drawn, so each link is made with both sides in view.

\paragraph{Boxes on the sketch.} Labels come from the COCO 80 categories extended with five that appear in these sketches but COCO does not name: tree, building, cloud, road sign, and tower. Every sketched object of one of these categories receives a box, tightened to the outer edge of its strokes. A box is drawn only when the strokes inside it are recognisable as that object on their own, without looking at the photograph. The ground truth is therefore self-contained.

\paragraph{Sketch-photo links.} The photograph's boxes are COCO's own annotations and stay untouched, so linking runs in one direction only. A sketched object is linked to a photographed instance when the two depict the same object, and the link is kept only when the sketch crop alone identifies that instance uniquely. An instance annotated in the photograph that also appears in the sketch receives a box on the sketch side and is then linked, while a sketched object with no counterpart among the photograph's annotations stays unlinked and serves the detection and pointing ground truth alone. A sketched object whose category COCO does not annotate therefore carries no query. This yields the 2754 query-target pairs of Sec.~\ref{sec:exp_datasets}.

\begin{figure}[!htbp]
\centering
\includegraphics[width=\textwidth]{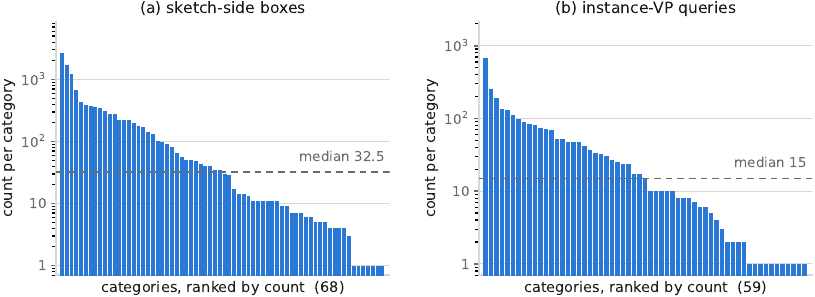}
\caption{\textbf{Per-category counts of the FS-COCO annotation.} Categories are ranked by count and the vertical axis is logarithmic. The dashed line marks the median.}
\label{fig:fscoco_stats}
\end{figure}

\paragraph{Category distribution.} The annotation holds 11674 boxes over 68 categories on 2907 sketches, 4.0 boxes per sketch, and the queries span 59 of those categories. Queries per category have a median of 15 against a mean of 47. The distribution is long-tailed (Fig.~\ref{fig:fscoco_stats}), as people drawing a scene from memory return to a few familiar objects far more often than to the rest.

\section{Affordance Evaluation Protocol}
\label{sec:supp_affordance}

AGD20K supplies a heatmap for each image and affordance verb, and its official metrics compare predicted and ground-truth heatmap distributions. Every model in Table~\ref{tab:affordance} emits points rather than a heatmap, so we score containment instead. The heatmap is binarised at half its peak value. Each connected region of the result becomes one ground-truth target, since one verb can apply to several instances in an image. Predicted points are matched to these targets one-to-one and greedily, as boxes are matched in Sec.~\ref{sec:exp_datasets}, and a match holds when the point falls inside its target. The label a model attaches to a point is ignored, so extra points count against precision. Recall and precision are computed over all matches, and F1 is their harmonic mean.

\section{Stratified Instance VP Across All Baselines}
\label{sec:supp_stratified}

Table~\ref{tab:stratified_supp} extends the stratification of Sec.~\ref{sec:exp_stratified} to every MLLM baseline. The split yields 689 Easy and 651 Hard queries on SketchyCOCO, and 2452 Easy and 302 Hard queries on FS-COCO. Base Rex-Omni performs near floor on this task, as Table~\ref{tab:instance_vp} shows, and is omitted. Every model loses F1 from Easy to Hard. Sketch-Omni leads every MLLM baseline on both Hard subsets and on FS-COCO-Easy, and is on par with Qwen3-VL-8B on SketchyCOCO-Easy.

\begin{table}[t]
\centering
\scriptsize
\caption{\textbf{Stratified instance VP, all baselines.} Each cell is SketchyCOCO\,/\,FS-COCO, same protocol as Table~\ref{tab:stratified}.}
\label{tab:stratified_supp}
\setlength{\tabcolsep}{5pt}
\renewcommand{\arraystretch}{1.1}
\begin{tabular}{ll|ccc}
\toprule
\textbf{Method} & \textbf{Subset} & \textbf{F1@0.3} & \textbf{F1@0.5} & \textbf{F1@0.7} \\
\midrule
Qwen2.5-VL-3B & Easy & 75.7/57.4 & 70.3/52.6 & 64.9/48.8 \\
 & Hard & 42.3/43.5 & 36.3/39.1 & 32.1/37.2 \\
Qwen3-VL-8B & Easy & \underline{91.0}/\underline{86.9} & \textbf{90.2}/\underline{84.3} & \textbf{86.7}/\underline{79.5} \\
 & Hard & \underline{50.5}/\underline{66.2} & \underline{47.8}/\underline{63.1} & \underline{43.4}/\underline{56.8} \\
MiMo-VL-7B & Easy & 83.9/51.9 & 82.9/48.9 & 79.0/43.0 \\
 & Hard & 47.3/30.5 & 43.8/28.9 & 39.8/26.1 \\
DeepSeek-VL2-S & Easy & 65.2/45.1 & 62.8/37.7 & 58.3/27.0 \\
 & Hard & 42.0/30.3 & 39.1/25.0 & 35.9/19.3 \\
\midrule
\textbf{Sketch-Omni} & Easy & \textbf{91.7}/\textbf{94.8} & \underline{89.5}/\textbf{92.7} & \underline{86.1}/\textbf{88.3} \\
 & Hard & \textbf{55.3}/\textbf{83.3} & \textbf{51.4}/\textbf{80.3} & \textbf{48.2}/\textbf{77.0} \\
\bottomrule
\end{tabular}
\end{table}

\section{Stage-A Alignment on the Sketch-Native Tasks}
\label{sec:supp_align_native}

Table~\ref{tab:align_native} evaluates the \emph{w/o align} configuration of Table~\ref{tab:instance_vp} on the five sketch-native tasks, isolating the alignment's effect on the tasks it operates on during Stage A. The largest gain falls on object referring, 4.3 points. This is consistent with the role of $\mathcal{L}_{\mathrm{cross}}$, which aligns how a phrase selects its region.

\begin{table}[t]
\centering
\scriptsize
\caption{\textbf{Stage-A alignment on the sketch-native tasks.} Detection and referring report F1@0.5, pointing F1@Point, affordance F1, all under the protocol of the main tables. Referring uses the test splits.}
\label{tab:align_native}
\setlength{\tabcolsep}{5pt}
\renewcommand{\arraystretch}{1.1}
\begin{tabular}{l|cc|cc|cc|cc}
\toprule
& \multicolumn{2}{c|}{\textbf{Detection}} & \multicolumn{2}{c|}{\textbf{Pointing}} & \multicolumn{2}{c|}{\textbf{Referring}} & \multicolumn{2}{c}{\textbf{Affordance}} \\
\textbf{Method} & FS-COCO & COCO & FS-COCO & COCO & obj & spat & Seen & Unseen \\
\midrule
\textbf{Sketch-Omni (full)} & 86.7 & 48.1 & 89.9 & 44.5 & 74.6 & 76.7 & 66.2 & 74.3 \\
\quad w/o align & 86.3 & 45.2 & 89.2 & 42.4 & 70.3 & 73.1 & 63.9 & 70.8 \\
\bottomrule
\end{tabular}
\end{table}

\section{Training Details}
\label{sec:supp_training}

\paragraph{Supervised stages.} Both stages fit the LoRA update $\Delta\theta$ with a single AdamW optimiser at an initial learning rate of $5{\times}10^{-5}$ with cosine decay, weight decay $0.01$, and gradient-norm clipping at $1.0$, at an effective batch size of $64$. The adapters use scaling factor $64$ and no dropout. Training runs in bfloat16 with gradient checkpointing under ZeRO stage 1, for about 1 day on 32 H100 GPUs.

\paragraph{Task mixture.} Stage A and Stage B alternate in blocks of 50 optimiser steps, two Stage-A blocks to every three Stage-B blocks. Within a block the tasks are sampled at the rates below, which give the share of all optimiser steps in the last column. Object and spatial referring share the referring stream and together receive probability 0.25.

\begin{center}
\begin{tabular}{llcc}
\toprule
\textbf{Stage} & \textbf{Task} & \textbf{Within stage} & \textbf{Overall} \\
\midrule
A & detection      & 0.25 & 0.10 \\
A & pointing       & 0.25 & 0.10 \\
A & referring      & 0.25 & 0.10 \\
A & affordance     & 0.25 & 0.10 \\
\midrule
B & instance VP, stitched canvases & 1/3 & 0.20 \\
B & instance VP, crop queries      & 1/3 & 0.20 \\
B & composited-scene detection     & 1/6 & 0.10 \\
B & composited-scene pointing      & 1/6 & 0.10 \\
\bottomrule
\end{tabular}
\end{center}

\paragraph{GRPO refinement.}\label{sec:supp_grpo} The final GRPO pass follows Rex-Omni's recipe and runs for about 8 hours on 32 H100 GPUs on the same task mixture as the supervised stages, with 8 rollouts per prompt at temperature 1.0 and top-$p$ 0.99, a global batch size of 32, learning rate $5{\times}10^{-6}$, PPO clip 0.2, and KL coefficient 0.01. Box tasks are rewarded by the IoU between the predicted and ground-truth boxes, and pointing tasks by whether the predicted point lands inside the ground-truth box or mask.

\end{document}